\documentclass[letterpaper, 10 pt, conference]{ieeeconf}  

\usepackage[final]{pdfpages}
\usepackage{amsfonts, amssymb, amsmath}
\usepackage{amsmath} 
\usepackage{url}
\IEEEoverridecommandlockouts                              

\usepackage{balance}

\newcommand{\qiyu}[1]{{\color{black}#1}}

\title{\LARGE \bf
GraspNeRF: Multiview-based 6-DoF Grasp Detection\\ for Transparent and Specular Objects Using Generalizable NeRF
}

\author{Qiyu Dai$^{1,2*}$, Yan Zhu$^{1*}$, Yiran Geng$^{1}$, Ciyu Ruan$^{3}$, Jiazhao Zhang$^{1,2}$, He Wang$^{1\dagger}$
\thanks{$^{1}$Center on Frontiers of Computing Studies, Peking University.}%
\thanks{$^{2}$Beijing Academy of Artificial Intelligence.}%
\thanks{$^{3}$College of Intelligence Science and Technology, National University of Defense Technology.}%
\thanks{$^{*}$The first two authors contributed equally.}%
\thanks{$^{\dagger}$Corresponding to hewang@pku.edu.cn.}%
}

\begin{document}

\maketitle
\thispagestyle{empty}
\pagestyle{empty}

\begin{abstract}
In this work, we tackle 6-DoF grasp detection for transparent and specular objects, which is an important yet challenging problem in vision-based robotic systems, due to the failure of depth cameras in sensing their geometry. 
We, for the first time, propose a multiview RGB-based 6-DoF grasp detection network, GraspNeRF, that leverages the generalizable neural radiance field (NeRF) to achieve material-agnostic object grasping in clutter. 
Compared to the existing NeRF-based 3-DoF grasp detection methods that rely on densely captured input images and time-consuming per-scene optimization, our system can perform zero-shot NeRF construction with sparse RGB inputs and reliably detect 6-DoF grasps, both in real-time. 
The proposed framework jointly learns generalizable NeRF and grasp detection in an end-to-end manner, optimizing the scene representation construction for the grasping.
For training data, we generate a large-scale photorealistic domain-randomized synthetic dataset of grasping in cluttered tabletop scenes that enables direct transfer to the real world.
Our extensive experiments in synthetic and real-world environments demonstrate that our method significantly outperforms all the baselines in all the experiments while remaining in real-time. \qiyu{Project page can be found at \url{https://pku-epic.github.io/GraspNeRF}.}

\end{abstract}

\section{INTRODUCTION}
Object grasping is a crucial technology for many robotic systems. 
Recent years have witnessed great progress in developing  vision-based grasping methods~\cite{fang2020graspnet,breyer2021volumetric,jiang2021synergies,sundermeyer2021contact,gou2021RGB,Wang_2021_ICCV,fang2022transcg}.
These methods rely on single-view or multiview depth inputs and can reliably detect 6-DoF grasps from cluttered table-top scenes mainly composed of objects with diffuse material. 
Despite the great progress in diffuse objects, grasping transparent and specular objects, which are ubiquitous in our daily life and need to be handled by robotic systems, is still very challenging.
Depth sensors struggle to sense those transparent and specular objects and usually generate wrong or even missing depths, thus further leading to the failure of grasping in those methods.

For grasping transparent and specular objects, recent works have been devoted to depth restoration and RGB-based grasping methods. 
Works, \textit{e.g.}, ClearGrasp~\cite{sajjan2020clear}, TransCG~\cite{fang2022transcg}, SwinDRNet~\cite{dai2022dreds}, leverage RGBD inputs and learn to recover the missing depths and correct the wrong depths as a separate step prior to performing grasping.
DexNeRF~\cite{ichnowski2022dex} provides a valuable alternative that only takes RGB inputs. It proposes to first construct a NeRF~\cite{mildenhall2020nerf}, the powerful implicit 3D representation that encodes scene geometry using an MLP, by capturing 49 RGB images from different viewpoints and then use volumetric rendering to render a top-view depth map that further passes to 3-DoF grasping module. Although being very novel, this method suffers from several important limitations. Its NeRF construction needs time-consuming per-scene training that takes at least  hours, which deviates far from the desired real-time running. This gets even worse for multi-object sequential grasping, since the scenes can change after each grasping and thus the NeRF needs to be retrained. A concurrent work, EvoNeRF~\cite{kerr2022evonerf} proposes to use Instant-NGP~\cite{mueller2022instant} to speed up NeRF training and evolve the NeRF after each grasp. However, it still costs 7s for each update, relies on dense image inputs, and can only perform 3-DoF grasping from top-down views. Another big disadvantage of using vanilla NeRF is that the scene geometry extracted from NeRF is usually imperfect and can be very low-quality for scenes composed of transparent objects and textureless backgrounds, given that NeRF is only overfitted to their RGB colors.
The wrong geometry can severely degrade the performance of the downstream pretrained grasping algorithms.

\begin{figure}[t]  
\begin{center}    
\includegraphics[width=\linewidth]{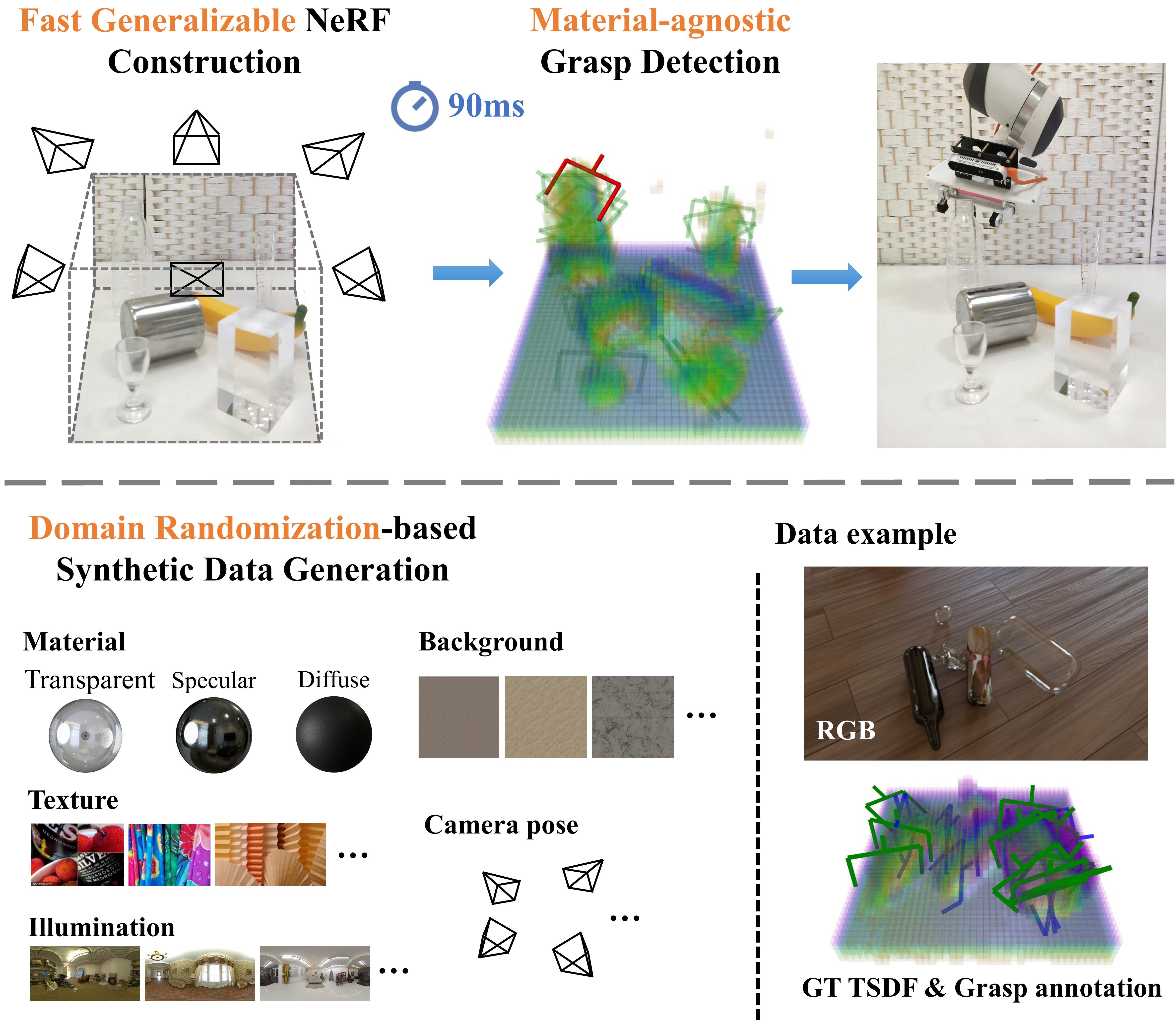}     \vspace{-6mm}   
\caption{Overview of the proposed GraspNeRF and the dataset. Our method takes sparse multiview RGB images as input, constructs a neural radiance field, and executes material-agnostic grasp detection within 90ms. We train the model on the proposed large-scale synthetic multiview grasping dataset generated by photorealistic rendering and domain randomization.} 
\vspace{-6mm}
\label{fig:teaser}  
\end{center}      
\end{figure}

To mitigate these issues, we propose to leverage generalizable NeRF, \textit{e.g.}, MVSNeRF~\cite{mvsnerf}, NeuRay~\cite{liu2022neuray}, which train on many different scenes and learn to aggregate multiview observations and zero-shot construct NeRFs for novel scenes without training. Moreover, it only requires sparse input views, which is more practical for real-world applications.
Our framework first constructs a 3D grid-like workspace and obtains the grid features via aggregating multi-view image features.
On these grid points, we then propose to predict truncated signed distance function (TSDF) to obtain a complete 3D geometric scene representation, which frees us from computing depths.
Taking input this TSDF, our volumetric grasp detection network can then predict 6-DoF grasps at each grid point. Note that this TSDF can also be turned into density and used for volume rendering.

To train this learning-based framework, we generate a large-scale photorealistic synthetic dataset of grasping diverse objects with diffuse, transparent, or specular material in cluttered table-top scenes. This dataset is composed of 2.4
million images with 100K scenes, rendered by physics-based rendering and enhanced by diverse domain randomization.
Trained on this extensive  dataset, our network can directly generalize to novel real-world scenes with novel objects.

Our experiments further show that the proposed method outperforms all baselines that leverage either depth restoration or separate NeRF construction, by over 20\% of grasp success rate, with 82.2\% and 65.9\% in packed and pile scenes, respectively on sequential transparent and specular object grasping in the real world. We also find an interesting phenomenon: although our grasping only cares about TSDF, enforcing NeRF rendering loss during training can bring performance improvements to grasping, especially on transparent and specular objects, indicating the synergies between view-dependent 2D rendering and grasping in 3D space.

To summarize, this proposed grasping algorithm has many remarkable advantages over the previous state-of-the-art works: it only takes sparse RGB inputs (we use 6) without depths; it is  material-agnostic, generalizable, and thus directly works on novel scenes; it is end-to-end differentiable and directly optimized for grasping;  it can perform 6-DoF grasping; and, finally, it runs at real-time with 11 FPS.




\section{RELATED WORK}
\subsection{Transparent and Specular Object Grasping}
Grasping transparent and specular objects is quite a challenging problem. Most depth-based methods fail due to incomplete depth maps. A direct way is to restore depths before detecting grasp. ClearGrasp\cite{sajjan2020clear} firstly estimates geometry information from a single RGB image and then refines depths via global optimization, which is used for 3-DoF transparent object grasping from top-down views. \cite{fang2022transcg} proposes a CNN-based depth completion network, and \cite{dai2022dreds} proposes a Swin Transformer-based RGBD fusion network, SwinDRNet, for depth restoration and the downstream 6-DoF grasping. 
Other studies focus on predicting grasp poses from RGB and depth.
\cite{weng2020multi} learns an RGBD grasp detection model that is transferred from an existing depth-based model.
\cite{cao2021fuzzy} predicts grasp poses from the RGBD-based CNN and executes grasping using the proposed soft robotic hand.
Those single-view-based grasping methods are difficult to handle object occlusions in cluttered scenes and cannot fully utilize the 3D geometry information.
Our method falls into the category which takes as input the multiview RGB images.
GlassLoc~\cite{zhou2019glassloc} constructs the Depth Likelihood Volume (DLV) descriptor from the multiview light field observations, representing the transparent object clutter scenes. 
GhostPose~\cite{chang2021ghostpose} conducts transparent object grasping by estimating the object 6D poses from the proposed model-free pose estimation approach based on multiview geometry. 
Departing from those methods, the proposed GraspNeRF represents the scene geometry as the generalizable NeRF, which is differentiable so that we are able to improve the grasping performance by jointly training with grasping and exploiting the synergies between both.

\subsection{Implicit Representations for Grasping}
Recent works show a trend in leveraging implicit scene representation for robotic tasks. GIGA~\cite{jiang2021synergies} proposes to leverage the synergies between geometry reconstruction and grasping prediction, which utilizes the convolutional occupancy network to construct the implicit scene representation. However, it takes a single-view depth image as input, which is hard to reconstruct transparent and specular objects.
\cite{yen2022nerfsupervision} renders depth maps from NeRF to provide dense correspondence for object descriptors applied to the manipulation task, which requires per-scene optimization and dense input. DexNeRF~\cite{ichnowski2022dex} is the first to utilize NeRF for grasping transparent objects, which avoids directly using the corrupted sensor depths by the proposed transparency-aware depth rendering. However, it is at the cost of capturing dense input images and hours of NeRF training for each grasp, which is impractical in real-world grasping. Furthermore, it only works on 3-DoF grasping, limiting its performance and applications.
The concurrent work, EvoNeRF~\cite{kerr2022evonerf}, speeds up the vanilla NeRF optimization using Instant-NGP~\cite{mueller2022instant},  
\qiyu{and MIRA~\cite{lin2022mira}, also utilizes Instant-NGP to synthesize a novel orthographic view, which is used to predict pixel-wise affordances for 6-DoF control.} However, they still suffer from dense inputs and require training before each grasp, while the proposed GraspNeRF leverages generalizable NeRF that is free from per-scene optimization and only needs sparse input, which is suitable for real-world grasping tasks.

\section{METHOD}

\begin{figure*}[htbp]  
\begin{center}    
\includegraphics[width=1.0\textwidth]{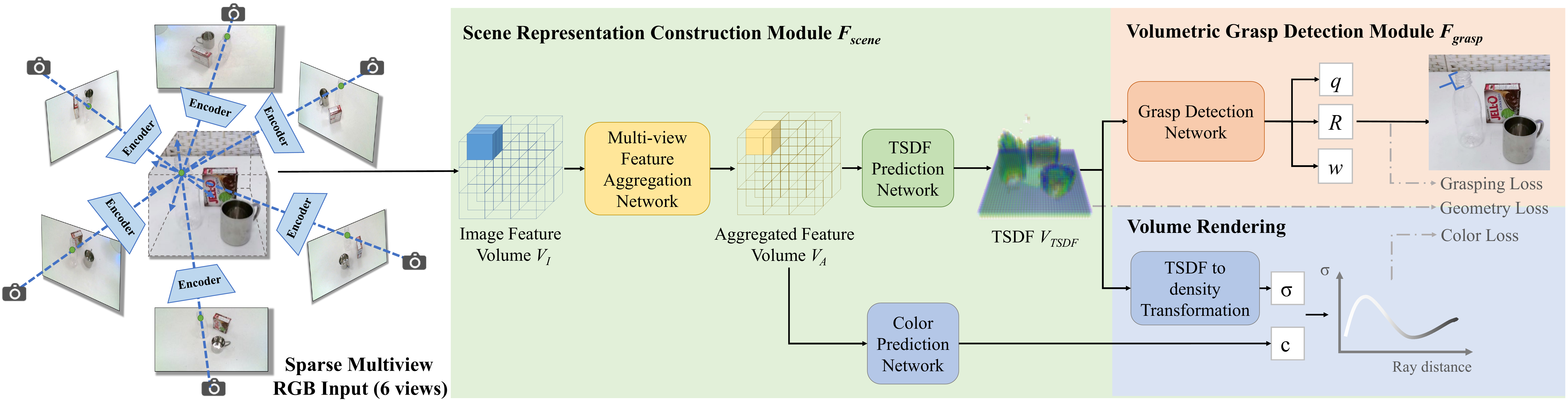}        
\caption{The framework of our proposed method. The features of input sparse multiview color images are first extracted to construct the image feature volume $V_{I}$ by querying the 2D image features corresponding to the 3D sample points. Then the implicit scene representation module maps $V_{I}$ to the predicted TSDF $V_{TSDF}$, pixel color ${c}$, and volume density $\sigma$. Volumetric grasp detection module $V_{TSDF}$ detects grasps with $V_{TSDF}$ as input.} 
\label{fig:pipeline}  
\vspace{-8mm}
\end{center}      
\end{figure*}

\subsection{Problem Statement and Method Overview}
Given sparse multiview RGB observations of a cluttered tabletop scene composed of transparent, specular, or diffuse objects, the goal of the proposed robotic system is to detect the 6-DoF grasps and then execute grasp-to-remove operations for all the objects. Here we assume the camera intrinsics and extrinsics are known for the multiview images.

We then formulate the 6-DoF grasp detection as a learning problem that maps a set of images from $N$ input views $\{ I_i \}_{i=1,...,N}$ to a set of 6-DoF grasps $\{ g_j | g_j = (t_j, R_j, w_j, q_j)\}$, where, for each detected grasp $g_j$, $t_j\in\mathbb{R}^3$ is 3D position, $R_j\in\mathrm{SO(3)}$ is 3D rotation, $w_j\in\mathbb{R}$ is opening width, and $q_j\in\mathrm\{0,1\}$ is the quality.

Our proposed framework is composed of a scene representation construction module $F_{scene}$ and a volumetric grasp detection module $F_{grasp}$, as shown in Figure~\ref{fig:pipeline}.
In $F_{scene}$, we first extract and aggregate the multiview image features for two proposes: the extracted geometric features form a feature grid and will be passed to our TSDF prediction network to construct the TSDF of the scene, which encodes the scene geometry as well as the density of the underlying NeRF; at the same time, the features are used to predict color, which along with the density outputs enables the NeRF rendering.
Taking the predicted TSDF as input, our volumetric grasp detection module $F_{grasp}$ then learns to predict 6-DoF grasps at each voxel of the TSDF.
\vspace{-1mm}
\subsection{Scene Representation Construction}
For our grasping framework, one crucial demand is to wholly and accurately represent the scene, especially for transparent and specular objects, in a generalizable and grasping-oriented way.
A vanilla NeRF needs scene-specific training and learns an MLP to overfit one scene without having an explicit representation.
We instead propose to leverage generalizable NeRF that learns to aggregate observations from multiple views and forms a volumetric feature grid. 

\textbf{Revisiting NeuRay}. 
The generalizable NeRF used in our scene representation construction module is inspired by NeuRay~\cite{liu2022neuray}, an occlusion-aware generalizable NeRF.
Taking as input a set of images $\left\{ I_i \right\}_{i=1,...,N}$, NeuRay first utilizes a shared CNN image encoder\qiyu{, which is trained from scratch} to extract the 2D image feature maps of $ C $ channels.
To render a pixel from a novel view, it samples $ K $ points along the pixel ray, projects them to each input view, and then queries image features, which forms $N$ ray features $ v_{I,i} \in \mathbb{R}^{ K \times C}  $.  To aggregate the feature from different views considering occlusion, NeuRay predicts a visibility weight $w_i$ indicating whether a 3D point is visible in each input view. With this weight, the ray features $\{v_{I,i}\}$ is aggregated to $v_A = \Sigma_i w_i v_{I,i}$.
Then it passes $v_A$ to a multi-head attention network for predicting the volume density of each sample point.
Volume colors are also predicted via blending pixel colors from input views. 

Note that the features in NeuRay are ray-based and do not form a volumetric representation of the whole scene, thus adopting NeuRay for grasping is non-trivial.

\textbf{Volumetric Feature Extraction}. To obtain grasping-oriented volumetric features, we first divide the 3D workspace into grids of size $S$. Each grid point $ x \in \mathbb{R}^{S^3}$ is projected to all of the $N$ views (N=6) and queries the corresponding feature, similar to NeuRay. This builds an image feature volume $ {V_{I}} \in \mathbb{R}^{S^3\times N \times C} $. Then, the image feature volume is aggregated among views using a multi-view feature aggregation network \qiyu{composed of MLPs}, forming the aggregated feature volume ${V_{A}} \in \mathbb{R}^{S^3 \times C}$. Now, for rendering, we could do the same to turn $V_A$ into color and density on the grid points. However, we argue that directly passing either $V_A$ or volume density to grasp detection network can be suboptimal, due to a lack of accurate geometric information.

\textbf{Truncated Signed Distance Field Prediction}. 
To obtain a scene representation that encodes accurate geometric information, we are inspired by NeuS~\cite{wang2021neus}, which predicts SDF instead of volume density and link these two together.
For grasp detection, we propose to predict TSDF from $V_A$. Specifically, we propose a TSDF prediction network that maps ${V_{A}}$ to the truncated signed distance field $ V_{TSDF} \in \mathbb{R}^{S^3}$. 
Inspired by NeuRay, for predicting the TSDF value at grid point $(u, v, w)$, we aggregate features from all $S$ grid points that share the same $u$ and $v$ using a multi-head attention network and the predicted TSDF value can be supervised by a geometry loss (see Section \ref{sec:loss}).

\textbf{Volume Rendering}. 
For volume rendering, we propose that this TSDF value can be tightly coupled with volume density $\sigma$ as shown below: 
\begin{equation}
    \sigma = \mathrm{max}\left(\frac{ -\frac{dS}{dx}{({F}_{TSDF}({x}))} }{S({F}_{TSDF}({x}))}, 0\right),
\end{equation}
where $S(\cdot)$ is the sigmoid function. In this way, since both grasp detection and volume rendering can leverage TSDF for their tasks, this enables the synergy between the two tasks. Note that the whole network can be learned in an end-to-end manner and directly optimized for grasping, whereas DexNeRF~\cite{ichnowski2022dex} needs \qiyu{staged training of NeRF and grasping}.

\subsection{Volumetric Grasping Detection}

The \qiyu{3D CNN} grasp detector $F_{grasp}$ learns a mapping from the TSDF grid to the dense grasp candidate volume, inspired by~\cite{breyer2021volumetric}. We consider the center of each voxel within the volume as the grasp position candidate. At given position $t \in \mathbb{R}^3$, the detector predicts three outputs from the separate decoder heads, including the grasping quality $q$, rotation $R$, and opening width $w$.

To select a proper grasp for execution, we discard the grasping position candidates far from the surface according to the predicted TSDF. Subsequently, we smooth the quality volume using Gaussian, filter out the grasp below the threshold and utilize the non-maximum suppression. We also choose the grasps that meet the gripper max opening width. 
Finally, we randomly select a grasp for robot execution from the remaining candidates \qiyu{to prevent repeated execution of an unsuccessful grasp due to a false positive.}

\subsection{Domain Randomization-based Synthetic Data Generation}
Existing multiview grasping datasets like~\cite{fang2020graspnet} do not consider transparent and specular objects, while the current multiview object datasets lack grasping annotations.
In this work, we present a large-scale synthetic multiview RGB-based grasping dataset, which contains 100K scenes with objects ranging from diffuse, specular to transparent, 2.4 million RGB images, and 2 million 6-DoF grasp poses. 

Our data generation pipeline is motivated by~\cite{dai2022dreds}. To construct the scenes, we transform an existing grasping dataset~\cite{jiang2021synergies} that contains CAD models and annotates object poses, by changing their object materials to diffuse, specular or transparent. Then we render photorealistic multiview RGBs on Blender~\cite{blender}.
To bridge the sim2real gap, we leverage domain randomization, which randomizes object materials and textures, backgrounds, illuminations, and camera poses. After training on the synthetic dataset with sufficient variations, the network considers real data as a variation of training data in testing time, so as to generalize to real.

\subsection{Network End-to-End Training}
\label{sec:loss}
We adopt end-to-end training for jointly learning scene representation and grasping, encouraging the two tasks to benefit each other. For grasping, it is crucial to perceive and reconstruct the scene geometry; and accurate grasping predictions, in turn, require high-quality scene representations. The training objective consists of three parts.

\textbf{Grasping Loss}. For grasping learning, we supervise the grasp quality, orientation, and opening width as done in~\cite{breyer2021volumetric}, which is formulated as:
\begin{equation}
   L_{grasp} = L_{q}(\hat{q},q)+q(L_{w}(\hat{w},w)+L_{o}(\hat{R},R)),
\end{equation}
where $L_{q}$ represents the binary cross-entropy loss between the predicted grasp quality label $\hat{q}\in\mathrm[0,1]$ and the ground truth $q\in\mathrm\{0,1\}$. $L_{w}$ is the $L_{2}$ loss penalizing width error. $L_{o}=1-|\hat{R} \cdot R|$ measures the quaternion distance. Note that because of the gripper's symmetry, a grasp rotated by $180^{\circ}$ around its wrist axis is deemed identical. Therefore, the orientation loss is determined by the closest distance from these two grasps to the ground truth. We also exclude the zero-quality grasp, encouraging the network to learn grasps with high confidence.

\textbf{Color Loss}. To learn a generalizable radiance field, 
we randomly take  6 training views as input for each training scene and randomly pick another targeted view for rendering. We supervise the novel view synthesis using $L_2$ loss:
\begin{equation}
   L_{color} = \sum_{k}{\Vert{\hat{C}_k - C_{k}}\Vert}_2,
\end{equation}
where $\hat{C_k}$ and $C_{k}$ denote the predicted and target color at pixel $k$ respectively.

\textbf{Geometry Loss}. Since the ground-truth TSDF values of each scene can be easily obtained from our synthetic data generation pipeline, we supervise the explicit geometry using $L_1$ loss along with the Eikonal regularization term~\cite{gropp2020implicit}:
\begin{equation}
\begin{aligned}
     L_{geo} = &\frac{1}{ \Vert \chi \Vert} \sum_{p_i \in \chi } ( \Vert \hat{V}_{TSDF}(p_{i}) - {V}_{TSDF}(p_{i}) \Vert_1 \\
   & + ({\Vert{\nabla{\hat{V}_{TSDF}(p_{i})}}\Vert}_2 - 1)^2 ),
\end{aligned}
\end{equation}
where $\hat{V}_{TSDF}(p_{i})$ and ${V}_{TSDF}(p_{i})$ is the predicted and ground truth TSDF values respectively. $p_{i}\in\chi$ denotes the 3D coordinate in the world frame.

The whole objective can be formulated as:
\begin{equation}
   L = L_{grasp} + L_{color} + L_{geo}.
\end{equation}

\section{EXPERIMENTS}
In this section, we evaluate the performance of our proposed method for grasping tasks through both simulation and real robot experiments. We also perform ablation studies on simulation data to analyze the impact of the main designs incorporated in our framework.

\subsection{Implementation Details}
The size of the TSDF volume is $40\times40\times40$. We train our network for 300k steps, utilizing the Adam optimizer with a learning rate of 1e-4 and a weight decay of 0.01. We randomly sample 2048 rays as a batch during training. The methods are evaluated on an NVIDIA RTX 3090 GPU.

\subsection{Experiment Setup}
\textbf{Grasping Environment Setup}. 
The Franka Emika Panda robot arm equipped with a parallel-jaw gripper performs grasping within a $30\times30\times30\,cm^3$ tabletop workspace. A RealSense D415 RGBD camera (depth is not used by our method) is attached to the gripper's wrist. To ensure adequate observation of objects, we uniformly sample 6 camera viewpoints on a hemisphere with a shared center in the workspace. All camera views point to the workspace center, represented in spherical coordinates as radius $r=0.5m$, polar angle $\theta=\frac{\pi}{6}$, and azimuthal angle $\varphi\in U(0,2\pi)$.

\textbf{Simulation Environment Setup}.
We build our simulation environment on PyBullet~\cite{coumans2021} and Blender~\cite{blender} for physically grasping simulation and multiview photorealistic image rendering, respectively. Note that synthetic depths obtained from the z-buffer cannot reflect the real depth missing and noise caused by specular and transparent materials. Thus we utilize the depth sensor simulator~\cite{dai2022dreds} to generate simulated depths with realistic sensor noise to serve as the baseline input. 

\textbf{Object Set}.
A total of 473 hand-scaled object meshes from~\cite{breyer2021volumetric} are used in simulation, with 417 for training, and 56 for testing. We randomize their textures and materials including transparent, specular, and diffuse. We also collect 20 household objects with various materials for real-world evaluations. In each scene, the objects are arranged as a \emph{pile} with random poses or \emph{packed} in an upright position.

\vspace{-1mm}
\subsection{Baseline Methods}
We compare our method with the following baselines:
\begin{itemize}
\item \textbf{VGN}~\cite{breyer2021volumetric}. A volumetric grasping detection network that takes the TSDF volume integrated from the multi-frame depths as input, and predicts the 6-DoF grasps.
\item \textbf{SwinDR-VGN}. A baseline that utilizes the SOTA depth restoration network, SwinDRNet~\cite{dai2022dreds}, to reduce depth errors from transparent and specular materials. It is designed to refine the TSDFs to benefit the VGN.
\item \textbf{NeRF-VGN}. A baseline that uses the SOTA NeRF method, Instant-NGP~\cite{mueller2022instant}, to learn the implicit scene representation and render depths from 49 camera viewpoints for TSDF integration. VGN then receives the TSDF and detects 6-DoF grasps. To make the approach perform appropriately, we give this baseline extensive viewpoints (49) compared with other baselines.
\end{itemize}

\begin{figure*}[htbp]  
\begin{center}    
\includegraphics[width=1.0\textwidth]{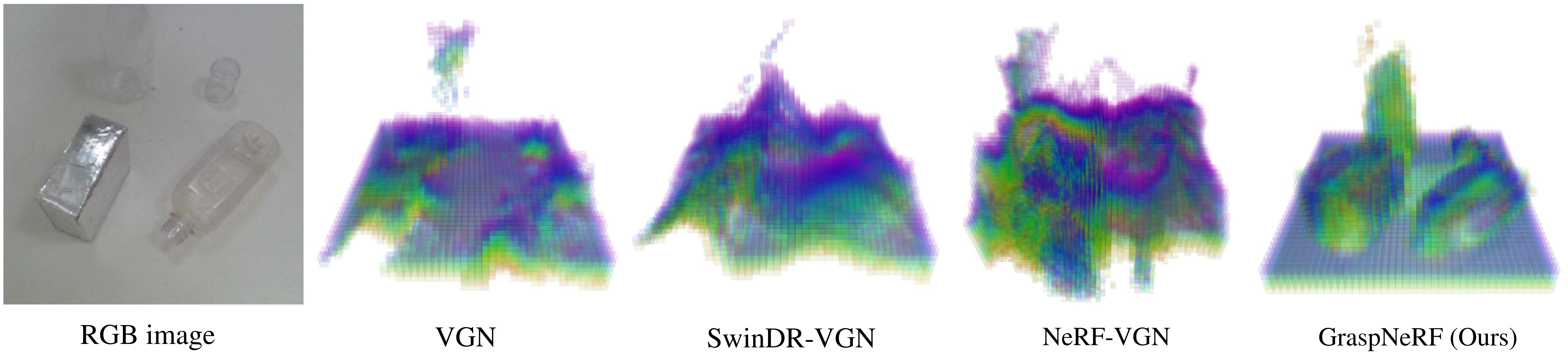}        
\caption{Comparison of TSDFs constructed by different methods. The geometry from VGN, directly fused from raw depth observations, is incomplete. Even restored by SwinDRNet, the geometry is still over-smooth and hard for grasp detection. NeRF-VGN also fails due to its inaccurate rendered depths under texture-less background. Only our method yields a good TSDF for specular and transparent objects, thanks to learning on diverse data.} 
\label{fig:geometry}  
\vspace{-8mm}
\end{center}      
\end{figure*}

\textbf{Evaluation Metrics}.
We measure the performance by 1) \textbf{Success Rate (SR)}: the ratio of the successful grasp number and the attempt number, and 2) \textbf{Declutter
Rate (DR)}: the average percentage of removed objects in all rounds.

\subsection{Simulation Grasping Experiments}
\textbf{Experimental Protocol}. In this subsection, we utilize extensive simulation data to evaluate all the methods. Our experiments include: 1) \textbf{Single object retrieval}: one object is placed at the center of the workspace. The robot arm performs grasps predicted by baselines, and a successful grasp indicates that the object is grasped out of the workspace. We evaluate with 12 test objects, 4 each of transparent, specular, or diffuse materials. Each object is tested three times, following the same protocol as EvoNeRF~\cite{kerr2022evonerf}, and the grasping success rate is reported.
2) \textbf{Sequential clutter removal}: we conduct 200 round table clearing experiments on both pile and packed scenes, each containing 5 objects. 
Each round shares the same objects and object poses, with two material combinations: A) transparent and specular, and B) a mix of all diffuse, transparent, and specular materials. 
In each trial of one round, the robot arm grasps and removes one object until the workspace is cleared or grasp detection fails or two consecutive failures are reached. Results are reported for both material combinations.

\textbf{Results and Analysis}. 
For single object retrieval, as presented in table~\ref{table:sim_single}, our method achieves an average success rate of 86.1\% over 36 trials, outperforming all the baselines. Furthermore, our method exhibits superior performance for sequential clutter removal, as shown in Table~\ref{table:sim_multi}, where we evaluate the methods on various objects with transparent and specular as well as mixed materials. Specifically, our method significantly improves the performance of transparent and specular object grasping compared to other methods.

For TSDF-based methods, the captured multiview depths contain sensor noises or missing values, especially on specular or transparent surfaces. This presents a challenge for maintaining multiview geometry consistency required by the TSDF, resulting in lower geometry quality that impacts the performance of VGN. SwinDR-VGN restores the depths before TSDF integration, thereby enhancing the TSDF quality to some extent. However, it suffers from over-smooth surfaces, causing objects to appear larger and potentially erasing smaller-sized objects in the TSDF, thus degrading the grasping performance.

For multiview RGB-based methods, NeRF-VGN requires 49 views and a long optimization time to overfit one scene, making it unsuitable for sequential clutter removal. Additionally, it exhibits a low grasping success rate in single object retrieval. 
We also observe that NeRF-VGN fails to perform adequately in textureless backgrounds, as illustrated in Figure~\ref{fig:geometry}, because vanilla NeRFs struggle to learn the geometry consistency of corresponding points in multiple views from poorly textured surfaces.

By comparison, our method, trained on a domain randomization-enhanced dataset that includes randomized backgrounds, learns a background prior during training, enabling it to generalize to both textureless and textured tabletop during inference. Figure~\ref{fig:geometry} demonstrates our superiority over competing methods for geometry reconstruction.

\begin{table}[h]
\caption{\qiyu{Success Rates (\%) of Single Object Retrieval in Simulation}}
\vspace{-3mm}
\begin{center}
\begin{tabular}{c|cccc}
\hline
 & Transparent & Specular & Diffuse & Overall \\
\hline
VGN & 25.0  & 41.7 & 91.7  & 52.8  \\
\hline
SwinDR-VGN & 58.3 & 66.7 & 91.7 &  72.2 \\
\hline
NeRF-VGN & 25.0 & 41.7 & 50.0 &  38.9 \\
\hline
\hline
Ours & \textbf{75.0} & \textbf{91.7} & \textbf{91.7} & \textbf{86.1} \\
\hline
\end{tabular}
\vspace{-3mm}
\end{center}
\label{table:sim_single}
\end{table}
\vspace{-4mm}

\begin{table}[h]
\caption{Results of Sequential Clutter Removal in Simulation\\Each cell: Transparent\&Specular
(Mixed)}
\vspace{-3mm}
\begin{center}
\begin{tabular}{c|cccc}
\hline
 & \multicolumn{2}{c}{Pile} & \multicolumn{2}{c}{Packed}\\
\hline
 & SR (\%) & DR (\%) & SR (\%) & DR (\%)\\
\hline
VGN & 40.6(48.9) & 26.0(33.7) & 56.0(61.8) & 57.3(64.0) \\
\hline
SwinDR-VGN & 57.9(61.7) & 40.6(42.1) & 43.8(54.8) & 36.0(49.2) \\
\hline\hline
Ours & \textbf{67.3}(\textbf{67.0}) & \textbf{46.2}(\textbf{45.8})& \textbf{82.3}(\textbf{83.8}) & \textbf{77.7}(\textbf{74.3}) \\
\hline
\end{tabular}
\vspace{-3mm}
\end{center}
\label{table:sim_multi}
\end{table}
\vspace{-3mm}

\subsection{Real Robot Experiments}
To evaluate the real-world performance of baselines, we further conduct real robot experiments.

\textbf{Experimental Protocol}. 
We follow the protocol of simulation experiments. For single object retrieval, we select 6 objects, including 2 each of transparent, specular, and diffuse, and place them in the same location with the same pose. The grasping iteration is repeated three times. For sequential clutter removal, we conduct 15 rounds of iteration on both pile and packed scenes, reporting results on transparent and specular object scenes, as well as mixed-material scenes. Each scene includes 4-6 randomly selected objects, and we capture RGB images from 6 views, as in the simulation environment. The network receives the multiview input and outputs 6-DoF grasp candidates, where the robot arm executes the top grasp. We consider a grasp successful if the object is removed from the workspace. The round ends when the grasp fails twice or all objects are cleared.


\textbf{Results and Analysis}. 
Table~\ref{table:real_single} reports the success rates of real-world single object grasping, where GraspNeRF obtains 88.9\% of 18 trials and beats competing baselines. For sequential clutter removal, GraspNeRF achieves the highest grasping success rate and declutter rate in various materials in both pile and packed settings, as shown in Table~\ref{table:real_multi}. 
\qiyu{Compared to NeRF-VGN (49 views), GraspNeRF only requires 6 views. Moreover, our method runs inference at 11 FPS without any optimization, which is significantly faster than NeRF-VGN with per-scene optimization. Importantly, our method successfully generalizes to new categories, as evidenced by the results for the two transparent objects (Glass gourd and small wine glass). These results demonstrate the effectiveness and robustness of our method in the real world. 

Nonetheless, we do note some failure cases.} For example, when grasping upright cylindrical objects in packed scenes, such as a hard glass bottle, the gripper might slip off the bottle due to insufficient friction.
\vspace{-1mm}

\begin{table}[h]
\caption{\qiyu{Success Rates (\%) of Real-world single object retrieval}}
\vspace{-3mm}
\begin{center}
\begin{tabular}{c|cccc}
\hline
 & Transparent & Specular & Diffuse & Overall\\
\hline
VGN & 33.3  & 66.6 & 100.0 & 66.6  \\
\hline
SwinDR-VGN & 50.0 & 50.0 & 83.3 & 61.1 \\
\hline
NeRF-VGN & 33.3 & 50.0 & 50.0 & 44.4 \\
\hline\hline
Ours & \textbf{83.3} & \textbf{83.3} & \textbf{100.0} & \textbf{88.9} \\
\hline
\end{tabular}
\label{table:real_single}
\end{center}
\end{table}
\vspace{-8mm}

\begin{table}[h]
\caption{Real-world Sequential Clutter Removal results\\Each cell: Transparent\&Specular(Mixed)}
\vspace{-3mm}
\begin{center}
\begin{tabular}{c|cccc}
\hline
 & \multicolumn{2}{c}{Pile} & \multicolumn{2}{c}{Packed}\\
\hline
 & SR (\%) & DR (\%) & SR (\%) & DR (\%)\\
\hline
VGN & 47.6(56.7) & 32.4(44.7) & 57.1(65.4)  & 49.3(59.7) \\
\hline
SwinDR-VGN & 44.6(57.3) & 29.7(39.5) &  64.3(71.2) & 53.7(66.3) \\
\hline
\hline
Ours & \textbf{65.9}(\textbf{72.4}) & \textbf{51.2}(\textbf{57.3}) & \textbf{82.2}(\textbf{81.6}) & \textbf{71.6}(\textbf{77.4}) \\
\hline
\end{tabular}
\vspace{-7mm}
\label{table:real_multi}
\end{center}
\end{table}

\subsection{Ablation Studies}

To analyze the design of GraspNeRF, we conduct ablation studies on various configurations. Experiments are conducted on the challenging pile scene in simulation under the sequential decluttering setting, which can be found in Table~\ref{table:ablation}.

We compare our end-to-end training strategy with two other schemes: 1) $F_{scene}$ and $F_{grasp}$ are trained separately, where $F_{grasp}$ is trained using perfect depth-integrated TSDF, and 2) train $F_{scene}$ first, then fix $F_{scene}$ to train $F_{grasp}$. The big performance gap (especially between ours and 1)) demonstrates the benefits brought by end-to-end training, which minimizes the impact of the TSDF errors on grasping.

We then analyze the effect of predicting TSDF on grasping detection. We let $F_{grasp}$ directly take the aggregated feature volume $V_A$ in $F_{scene}$ and thus remove the geometry supervision. Compared to TSDF input, the performance significantly degrades, demonstrating that supervised-learned geometric representation is essential for volumetric grasping learning.

To further analyze the synergies between neural radiance field and grasping, the volume rendering branch of $F_{scene}$ and its color loss are removed. We observe that the performance drops, especially on transparent and specular object declutter rate, indicating that image supervision helps grasp detection on transparent and specular objects.
\vspace{-2mm}


\begin{table}[h]
\caption{Ablation Studies on Sequential Pile Object Removal\\Each cell: Transparent\&Specular(Mixed)}
\label{table:ablation}
\vspace{-3mm}
\begin{center}
\begin{tabular}{c|cccc}
\hline
 & SR (\%) & DR (\%)\\
\hline
Train $F_{scene}$ \& $F_{grasp}$ separately & 45.0(47.5) & 32.5(33.9)\\
\hline
Fix $F_{scene}$, train $F_{grasp}$  & 65.0(64.0) & 41.1(40.2) \\
\hline
w/o TSDF & 57.8(55.9) & 37.5(36.0)\\
\hline
w/o volume rendering & 66.0(65.9) & 42.1(44.0)\\

\hline
\hline
Full model & \textbf{67.3}(\textbf{67.0}) & \textbf{46.2}(\textbf{45.8}) \\
\hline
\end{tabular}
\end{center}
\end{table}
\vspace{-6mm}

\section{CONCLUSIONS}
In this work, we propose GraspNeRF, a multiview RGB-based 6-DoF grasp detection network for transparent and specular objects. Our method extends the generalizable NeRF to grasping, enabling sparse input and direct real-time inference without optimization. We propose the joint learning of scene representation and grasping to exploit the synergy between them. We also present a large-scale synthetic multiview grasping dataset, leveraging domain randomization to bridge the sim-to-real gap. Experiments demonstrate our superiority over competing methods. \qiyu{One limitation lies in the volume grid, whose limited resolution may be the bottleneck to the performance of geometry reconstruction and grasping. We leave further improvements to future works.}









\bibliographystyle{IEEEtran}
\balance
\bibliography{IEEEabrv,reference}

\end{document}